# Incremental Dynamic Construction of Layered Polytree Networks

**Keung-Chi Ng**
IET Inc.,
14 Research Way, Suite 3
E. Setauket, NY 11733
kcng@iet.com

**Tod S. Levitt**
IET Inc.,
14 Research Way, Suite 3
E. Setauket, NY 11733
levitt@iet.com

## Abstract

Certain classes of problems, including perceptual data understanding, robotics, discovery, and learning, can be represented as incremental, dynamically constructed belief networks. These automatically constructed networks can be dynamically extended and modified as evidence of new individuals becomes available. The main result of this paper is the incremental extension of the singly connected polytree network in such a way that the network retains its singly connected polytree structure after the changes. The algorithm is deterministic and is guaranteed to have a complexity of single node addition that is at most of order proportional to the number of nodes (or size) of the network. Additional speed-up can be achieved by maintaining the path information. Despite its incremental and dynamic nature, the algorithm can also be used for probabilistic inference in belief networks in a fashion similar to other exact inference algorithms.

**Keywords**: incremental, dynamically constructed networks, incremental inference, polytree, layered clustering, layered belief networks, node aggregation

## 1 INTRODUCTION

Belief networks are directed acyclic graphs that represent and manipulate probabilistic knowledge (Neapolitan 1990, Pearl 1988). In a belief network, the nodes represent sets of random variables and the arcs specify their conditional dependence. The compact representation, sound theory, and the existence of inference algorithms have made belief networks a popular knowledge representation for uncertain reasoning in expert systems.

Many exact probabilistic inference algorithms[1] have been developed and refined. Among the earlier methods, the polytree algorithm (Kim and Pearl 1983, Pearl 1986) can efficiently update singly connected belief networks (or polytrees), however, it does not work (without modification) on multiply connected networks. In a singly connected belief network, there is at most one path (in the undirected sense) between any pair of nodes; in a multiply connected belief network, on the other hand, there is at least one pair of nodes that has more than one path between them. Despite the fact that the general updating problem for multiply connected networks is NP-hard (Cooper 1990), many propagation algorithms have been applied successfully to multiply connected networks, especially on networks that are sparsely connected. These methods include clustering (also known as node aggregation) (Chang and Fung 1989, Pearl 1988), node elimination and arc reversal (Shachter 1986), conditioning (Pearl 1986), revised polytree algorithm with cutset conditioning (Peot and Shachter 1991), graph triangulation and clique-tree propagation (Lauritzen and Spiegelhalter 1988), the join-tree approach (Jensen et. al. 1990), and symbolic probabilistic inference (Shachter et. al. 1990). Despite their successes, these methods have only been applied to "static" belief networks. A static network has fixed structure and only allows changes in the instantiation of evidence. Furthermore, these methods are not sufficient for incremental extensions of belief networks that are an obvious approach in certain class of problems, such as perception, robotics, discovery, and learning.

In this paper, we develop an algorithm that can construct a layered polytree network incrementally and dynamically, in such a way that the resulting network is also a polytree. The keys to the algorithm are the layered characteristics of the network and the removal of cycles by clustering nodes, without changing the underlying joint distributions of the network. After clustering, the reduced network becomes singly connected and the distributed polytree algorithm is applied to the reduced network. This approach, called *layered clustering*, is very similar to the node aggregation proposed by Chang and Fung (Chang and Fung 1989) however, the algorithm is deterministic and does not require search. Moreover, the algorithm is inherently incremental and can be used for incremental and dynamic extensions to belief networks.

---

[1] Due to the computational complexity of the exact probabilistic inference, many approximate inference methods have been proposed, such as (Chavez and Cooper 1990, Chin and Cooper 1987, Fung and Chang 1989, Henrion 1988, Pearl 1987, Shachter and Peot 1990). Approximate probabilistic inference, which is also NP-hard (Dagum and Luby 1993), is not the focus of this paper.

Section 2 defines layered belief networks and shows how to convert any belief network into a layered one. Section 3 details how to construct a layered polytree network incrementally and dynamically, and describes the layered clustering algorithm and illustrates it with some examples. Section 4 outlines how the algorithm can be used for dynamic extension in belief networks and outlines an incremental polytree algorithm. Finally, the algorithm's computational complexity is discussed.

## 2 LAYERED BELIEF NETWORKS

In a belief network, we can assign a *level* (or *depth*) to each node $X$ to denote the number of nodes in its longest unidirectional path between $X$ and all root nodes. For example, in Figure 1, root nodes $A$, $B$, and $C$ are at level 0 (the lowest level), nodes $D, E$, and $F$ are at level 1, nodes $G$ and $H$ is at level 2, and nodes $I$ and $J$ are at level 3. (We use $A, B, C$, etc., to represent a node or set of random variables, and use $A_i$ and $B_j$ to denote the $i$th value of the joint set of states of the set $A$ and the $j$th value of the joint set of states of the set of random variables in $B$ respectively.)

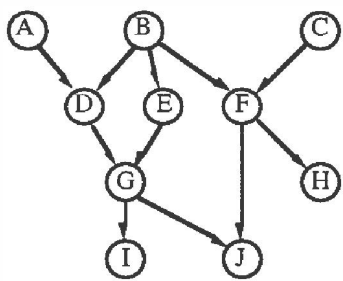

Figure 1: An example of a belief network.

A *layered* belief network is a belief network in which all the direct predecessors (or parents) of each node $X$ at level $i$ are at level $i-1$ and all its direct descendants (or children) are at level $i+1$. It can be observed that the belief network in Figure 1 is not a layered belief network (e.g., $F$ of level 1 has a direct descendant $J$ at level 3). It is, however, very easy to convert a belief network into a layered belief network with the addition of "intermediate" nodes. For example, Figure 2 shows a layered belief network for the network shown in Figure 1, with the addition of a new node $F'$.

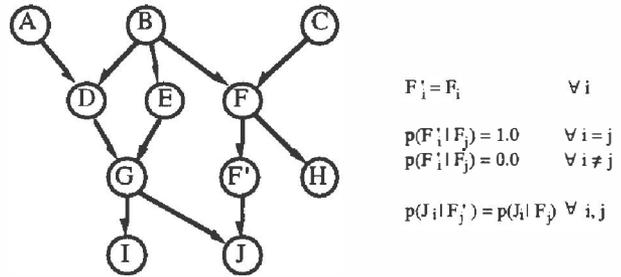

Figure 2: A layered belief network for the one shown in Figure 1.

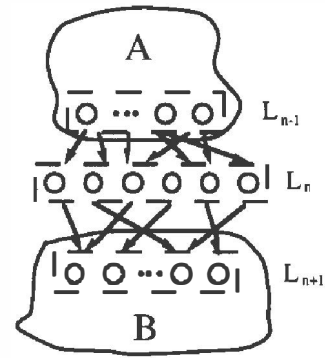

Figure 3: In a layered belief network, every path between $A$ and $B$ must pass through $L_n$.

In a layered belief network, it is trivial to find sets of nodes that separate or decompose the network into two parts. One such set is the collection of all the nodes at level $n$, denoted as $L_n$. In the network shown in Figure 3, the nodes in $A$ and $B$ are connected to one another through the nodes in $L_n$. The structure of the network also implies that $B$ is conditionally independent of $A$ given $L_n$, or $L_n$ *d-separates* (Geiger et. al. 1990) $B$ from $A$. More formally,

$$p(B|L_n, A) = p(B|L_n).$$

Consider the network shown in Figure 3 that consists of a set of nodes at level $n$, labeled $L_n$, with the set of all its parents, $L_{n-1}$, and all its children, $L_{n+1}$. Let $e$ denotes the total evidence, where $e_A$ is the evidence connected to $L_n$ through its parents $L_{n-1}$, and $e_B$ is the evidence connected to $L_n$ through its children $L_{n+1}$. We have,

$$BEL(L_n) = p(L_n|e) = \alpha \, p(e_B|L_n) \, p(L_n|e_A),$$

where $\alpha$ is a normalization constant. Denoting $\pi(L_n) = p(L_n|e_A)$ and $\lambda(L_n) = p(e_B|L_n)$, the above equation becomes

$$p(L_n|e) = \alpha \, \pi(L_n) \, \lambda(L_n).$$

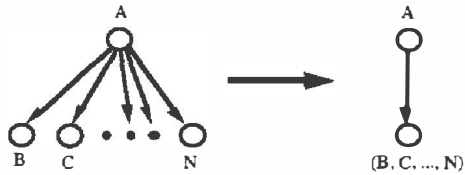

Figure 4: An example that illustrates poor node clustering.

## 3 INCREMENTAL DYNAMIC CONSTRUCTION OF POLYTREES

Before we can apply the layered clustering algorithm to construct a polytree incrementally and dynamically, we have to obtain the level information for all the nodes in a belief network. The level for a node $X$ can be determined recursively as follows:

```
if node X has no parent then begin
    level(X) = 0;
end else begin {node X has some parent(s)}
    level(X) = 1 + max[level(parent_1),
                        level(parent_2), ...];
end; {if}
```

Based on the level information, we can determine if we need to convert the network to a layered belief network with the addition of intermediate nodes. Once a layered belief network is formed, we can apply the layered clustering algorithm to the network.

### 3.1 LAYERED CLUSTERING ALGORITHM

A very straight forward (but naive) approach to layered clustering is to aggregate all the nodes at a level into a compound node and the resulting polytree will degenerate into a chain. This approach does not make full use of the structure and the independence relationships among the nodes. For example, this algorithm clusters all the nodes $B, C, \ldots, N$ in the network shown in Figure 4 into a single compound node, although they are all singly connected through the node $A$ and are conditionally independent given $A$. A good layered clustering algorithm uses the structure of the network, such as the presence of undirected cycles, to determine which nodes to cluster.

Given a layered belief network that is connected (i.e., there is a path between any two nodes in the network), nodes can be clustered incrementally by:

1. initialize an empty network $K$,
2. pick a node $V$ in the original network, and
3. call AddNode($V$).

The procedure AddNode($V$) is defined recursively as follows:

```
Add node V to K and mark V as ADDED;
while there are more arcs to add do begin
    Add an arc that links V to a node in K;
        {a cycle is formed}
    Find cycle C;
    LayerCluster(C);
end; {while}
U = first parent of V;
while U is not NULL do begin
    if U is not marked as ADDED then begin
        Add arc U → V to K;
        AddNode(U);
    end; {if}
    U = next parent of V;
end; {while}
W = first child of V;
while W is not NULL do begin
    if W is not marked as ADDED then begin
        Add arc V → W to K;
        AddNode(W);
    end; {if}
    W = next child of V;
end; {while}
```

There are a number of standard deterministic algorithms (see Aho et. al. 1974) that can be used to find the cycle $C$. Once the cycle is found, the LayerCluster($C$) procedure is very straight forward. First, we find all the sink nodes (i.e., nodes that do not have any descendants) in $C$. For each of these sink nodes $U$, we combine the parents of $U$ (there are exactly 2) into an aggregated node. This combination process is repeated for all the aggregated nodes until all the nodes in $C$ have been processed.

Figure 5 shows the incremental steps when the layered clustering algorithm is applied to the network as shown in Figure 2 (we actually start with node $A$, a root node). Several more examples of the algorithm can be found in Figure 6.

### 3.2 DISCUSSION

The way in which nodes and arcs are added is very similar to a depth-first traversal. In this manner, we ensure that the resulting network is connected. If we start with a connected polytree, the addition of a new node and one arc to the polytree results in another connected polytree. However, the addition of one extra arc between any two nodes (say $X$ and $Y$) in a connected polytree introduces a cycle. This is because there is already one path between $X$ and $Y$ in the polytree and the new arc becomes another path. Because we know how a cycle can be introduced in a connected polytree, there is no need for cycle detection in the AddNode procedure.

Suppose that a new node $V$ is added with $k$ arc connections to nodes in the polytree (assuming that the addition of $V$ with $k$ arcs does not violate the layered requirement for the algorithm), then there can be at most $_kC_2$ cycles (i.e., the number of cycles is equal to the number of all possible ways to get pairs of 2 nodes out of $k$ nodes). Because layered clustering eliminates a cycle once it is detected, we only have to repeat the cycle elimination step $(k-1)$ times. In the steps shown in Figure 5, all the nodes in the network have their levels precomputed. When the level of node $V$ is unknown, then it is necessary to check whether the $k$ arc connections create "illegal" cycles (e.g., cycles that violate the acyclic property of a belief network). If all the $k$ arcs are either going from $V$ to the polytree or vice versa, the addition of $V$ and the $k$ arcs does not create any illegal cycles, but it may still be necessary to add intermediate nodes to retain the layered polytree structure. On the other hand, when some of the $k$ arcs are from $V$ to the polytree and the rest from the polytree to $V$, then it is necessary to find out whether there are any "back" arcs (i.e., arcs that go from a node at a higher level to $V$ and then to a node at a lower level). To do this, we first find out the minimum level the nodes for all the arcs that go from $V$ to the polytree (labeled as out(min)) and the maximum level of the nodes for all arcs that go from the polytree to $V$ (labeled as in(max)). If out(min) is larger than in(max), then we can add node $V$ and its $k$ arcs. If in(max) is larger than out(min), then we should not allow the addition because back arcs are created and the layered property of the polytree is violated. Actually, if the difference in level between in(max) and out(min) is 1, the addition of node $V$ leads to the creation of more intermediate nodes and an extra level between in(max) and out(min), as well as the necessary intermediate nodes caused by the addition of $V$ to maintain the layered polytree.

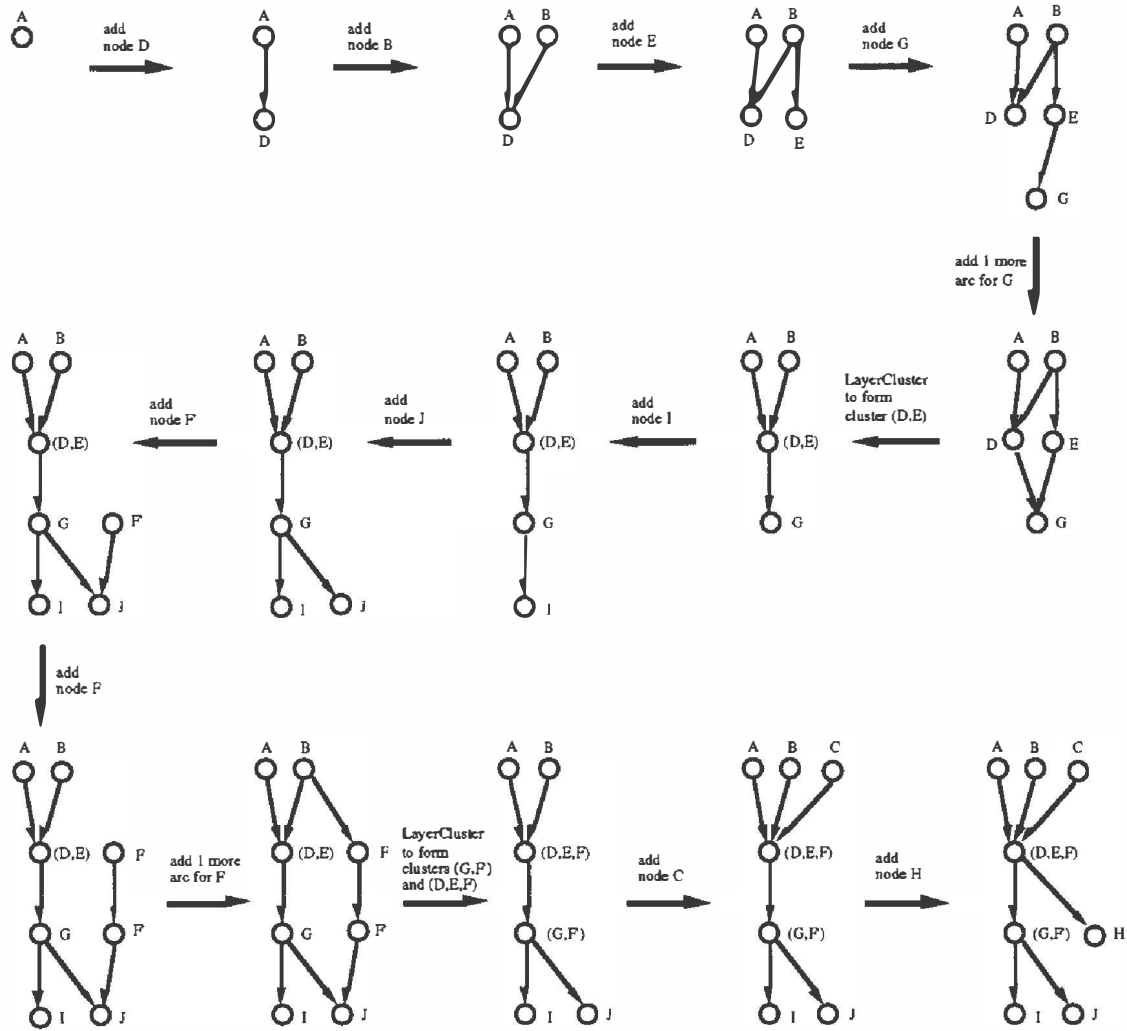

Figure 5: An example that illustrates the steps of the incremental, dynamic construction of a polytree network with layered clustering for the network shown in Figure 2.

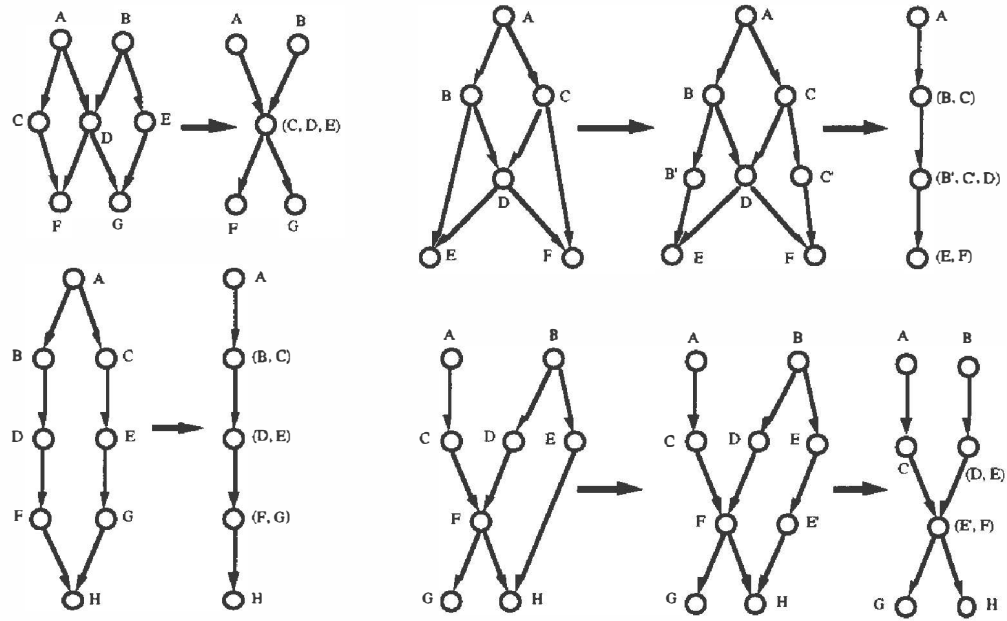

Figure 6: More examples on the layered clustering algorithm.

The complexity of the AddNode procedure is $O(l+v)$, where $l$ is the total number of arc and $v$ is the total number of nodes for the layered network. Because the smallest cycle that can form in a layered network requires 4 nodes and 4 arcs (see Figure 7(a)) and it requires at least 2 more arcs and a new node to create another cycle (see Figure 7(b)), at most there can be $(\frac{l}{2}-1)$ cycles in a layered network. Standard algorithms for finding a cycle are $O(v)$ and it takes $O(v)$ for the LayerCluster procedure. Thus, the complexity for the whole layered clustering algorithm is $O(l\,v)$.

Although not proven, we believe that the layered clustering produces close to the optimal clustering of nodes in a multiply connected layered network. Once the multiply connected network is converted to a polytree, we can apply the polytree algorithm to compute the belief for the variables.

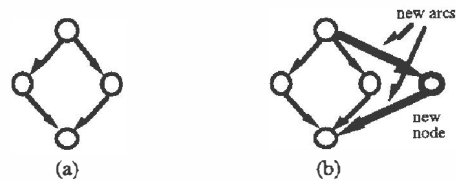

Figure 7: Figure showing the smallest cycle in a layered network and that at least 2 new arcs and 1 new node is needed to create an additional cycle.

Depending on the network structure, some inference methods may perform better than others (Suermondt and Cooper 1991). In order to apply the layered clustering, the network has to be layered. There are many network structures, such as a fully connected network that requires the addition of a significant number of intermediate nodes in order to convert them to layered networks. Because the polytree algorithm has a complexity of $O(\Gamma^2)$, where $\Gamma$ is the maximum state space size for any node (simple or cluster one) in the network, the addition of many intermediate nodes may increase the state space size for the cluster nodes. For such networks, it is not clear whether layered clustering is as efficient as other inference methods.

Apart from being an efficient clustering algorithm with layered networks, the layered clustering does not rely on heuristics on the choice of nodes to cluster. Thus, it results in the same clustering regardless of which node is chosen as the initial node. Another advantage is that it can be implemented incrementally (see Figure 5 for an illustration). Thus, it can be used for dynamic extension of belief networks. However, in order to fully apply layered clustering for incremental inference, the polytree algorithm must be modified accordingly.

## 4 INCREMENTAL POLYTREE ALGORITHM

Incremental changes (mainly addition of arcs and/or nodes) to a polytree can be:

1. changes that do not introduce a cycle, such as the addition of a new node and *only* one arc (see Figure 8(a)), or

2. changes that introduce a cycle to the polytree, such as the addition of an arc between 2 nodes in the polytree (see Figure 8(b)).

When there is no cycle being created by the addition of a new node and arc, the polytree algorithm can incorporate the impact of the newly added node $V$ as follows:

- If $V$ is a root node, then $V$ sends a message $\pi(V)$ to its only descendant $W$ and $W$ sends a $\lambda(W)$ message to $V$ so that $V$ can compute its new belief. Also, the link matrix (i.e., conditional probabilities) of $W$ has to be updated.

- If $V$ is a leaf node, then $V$ sends a message $\lambda(V)$ to its only predecessor $U$ and $U$ sends a $\pi(U)$ message to $V$ to calculate the new belief.

In case where a cycle is created, the nodes in the cycle are first clustered according to the layered clustering algorithm. Then, the link matrices for all the nodes that get combined are updated. For example, in Figure 5, after forming a cluster $(G, F')$, we compute the new conditionals with the formula $p(G, F'|D, E, F) = p(G|D, E) * p(F'|F)$. After that, the root nodes of the cycle (after clustering) send $\pi$ messages and the leaf nodes send $\lambda$ messages to their neighbors. In the example shown in Figure 5, if we consider the cycle $B, (D, E), G, J, F', F$, then $B$ sends message $\pi(B)$ and $J$ sends message $\lambda(J)$. At the same time, the nodes that are originally connected directly to the cycle (i.e., $A$ and $I$ from the same example) send the appropriate messages, or $\pi(A)$ and $\lambda(I)$ messages to their neighbors. These $\pi$ and $\lambda$ messages propagate through the polytree and the beliefs of all nodes get updated.

There are situations in which the added node is not a true root node, but a "pseudo" root node, i.e., the node is actually not a root node in the original network, but the way that it is added to the polytree resembles that of a root node (e.g., the addition of $F'$ in Figure 5). There is no way for the pseudo root node to generate the $\pi$ message because the information about its direct predecessor is missing. The $\pi$ message is sent once such information becomes available (e.g., after the addition of more root nodes to the pseudo root node).

## 5 CONCLUSION

We have detailed an algorithm that constructs a layered polytree incrementally and dynamically. The algorithm is very efficient and it can be used for probabilistic inference in multiply connected layered belief networks. The algorithm is deterministic and does not require any search or heuristics to determine the nodes to be clustered. The complexity of adding a new node with a cycle is at most $O(v)$, where $v$ is the number of nodes in the network. A method to convert any belief network to become a layered belief network is outlined and a modified polytree algorithm is also presented.

The incremental algorithm is applicable for problems in perception and robotics where a polytree model suffices. In addition, the algorithm is applicable to other problems that can be modeled as belief networks but require incremental extensions, as well as most other problems in which belief networks have been used (such as diagnostic domains). The layered clustering algorithm can be used for probabilistic inference in the same fashion as the other exact inference algorithms and is a promising inference method for layered network, both sparsely and highly connected ones. For layered networks that are highly connected (such as networks used in computer vision (Agosta 1991, Levitt et. al. 1989)), we speculate that this approach is more efficient than the undirected cliques method. This is because the clique methods generates a large number of highly interrelated cliques while there is little (or no need) for the introduction of intermediate nodes with layered clustering. The incremental nature of layered clustering and the polytree algorithm is very suitable for incremental inference. More detailed study and comparison between these inference methods is an ongoing and promising area of research.

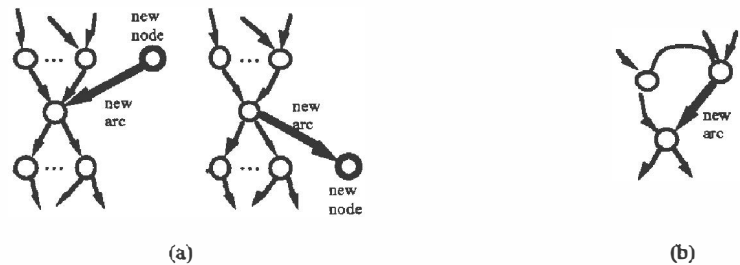

Figure 8: The two ways that incremental changes are made to a polytree.

## Acknowledgments

This work is partly supported by government contract DACA76-93-C-0025.